%% file: sample-sigconf.tex
\documentclass[sigconf]{acmart}

\usepackage{booktabs} %
\usepackage{cleveref}
\usepackage{algorithm}
\usepackage{algpseudocode}
\usepackage{amsmath}
\usepackage{tcolorbox} 
\usepackage{multirow}

\setcopyright{acmcopyright}

\acmDOI{https://doi.org/10.1145/3672608.3707810
}

\acmISBN{979-8-4007-0629-5/25/03}

\acmConference[SAC'25]{ACM SAC Conference}{March 31 –April 4, 2025}{Sicily, Italy}
\acmYear{2025}
\copyrightyear{2025}

\acmArticle{4}
\acmPrice{15.00}

\begin{document}
\title{Curriculum Demonstration Selection for In-Context Learning}
  
\renewcommand{\shorttitle}{SIG Proceedings Paper in LaTeX Format}

\author{Duc Anh Vu}
\affiliation{%
  \institution{Nanyang Technological University}
  \country{Singapore}}
\email{vuducanh001@e.ntu.edu.sg}

\author{Cong-Duy Nguyen}
\affiliation{%
  \institution{Nanyang Technological University}
  \country{Singapore}}
\email{nguyentr003@ntu.edu.sg	}

\author{Xiaobao Wu}
\affiliation{%
  \institution{Nanyang Technological University}
  \country{Singapore}}
\email{xiaobao002@e.ntu.edu.sg}

\author{Hoang Minh Nhat}
\affiliation{%
  \institution{Nanyang Technological University}
  \country{Singapore}}
\email{mnhat.hoang2002@gmail.com	}

\author{Du Mingzhe}
\affiliation{%
  \institution{Nanyang Technological University}
  \country{Singapore}}
\email{mingzhe001@e.ntu.edu.sg	}

\author{Nguyen Thanh Thong}
\affiliation{%
  \institution{National University of Singapore}
  \country{Singapore}}
\email{e0998147@u.nus.edu	}

\author{Anh Tuan Luu}
\authornote{Corresponding author.}
\affiliation{%
  \institution{Nanyang Technological University}
  \country{Singapore}}
\email{anhtuan.luu@ntu.edu.sg}

\renewcommand{\shortauthors}{Anh et al.}

\begin{abstract}

Large Language Models (LLMs) have shown strong in-context learning (ICL) abilities with a few demonstrations.
However, one critical challenge is how to select demonstrations to elicit the full potential of LLMs.
In this paper, we propose Curriculum Demonstration Selection (CDS), a novel demonstration selection method for ICL.
Instead of merely using similarity, CDS additionally partitions samples by their complexity measurements. 
Following curriculum learning, CDS then selects demonstrations from easy to difficult.
Thus the selected demonstrations cover a wide range of difficulty levels, enabling LLMs to learn from varied complexities within the training set.
Experiments demonstrate that our CDS consistently outperforms baseline methods, achieving notable improvements across nine LLMs on three benchmarks. Moreover, CDS proves especially effective in enhancing LLM performance in solving challenging problems.

\end{abstract}



\maketitle

\input{samplebody-conf}

\bibliographystyle{ACM-Reference-Format}
\bibliography{sample-bibliography} 
\input{appendiz}

\end{document}

%% file: samplebody-conf.tex
\section{Introduction}

Large language models (LLMs) have made significant strides in natural language processing (NLP) \cite{pmlr-v119-zhang20ae, touvron2023llama, penedo2023refinedweb}, excelling in tasks such as natural language understanding \cite{hendrycks2020measuring, brown2020language, hoang-etal-2024-toxcl, wu2024akew}, text generation \cite{10.1145/3605098.3636030, zhang2024benchmarking,nguyen2021enriching,nguyen2022improving} and reasoning \cite{kojima2022large, qiao2022reasoning,pan2023fact,pan2024fallacy}. Despite these advances, the full potential of LLMs remains limited by how they generalize across tasks of varying complexity. As LLMs tackle increasingly diverse and complex real-world challenges, optimizing their ability to handle varying task difficulties is crucial.

In-Context Learning (ICL) has emerged as a powerful paradigm in which LLMs perform tasks by leveraging examples embedded within the input prompt, without the need for parameter updates. The success of ICL, however, is highly sensitive to the choice of demonstrations included in the prompt. Studies have shown that selecting the right set of demonstrations can significantly influence model performance \cite{liu-etal-2022-makes, dong2022survey}. Despite this, existing methods for demonstration selection often rely on random selection or heuristic approaches that may lead to inconsistent and suboptimal outcomes, particularly in more complex tasks.

Although various approaches have been proposed to improve demonstration selection, including unsupervised \cite{sorensen-etal-2022-information, liu-etal-2022-makes, wu-etal-2023-self,nguyen2021contrastive} and supervised methods \cite{rubin-etal-2022-learning, zhang-etal-2022-active, wang2024large,nguyen2025meta}, these techniques often fail to account for the varying difficulty of the demonstrations. As a result, LLMs are limited in their ability to generalize effectively across tasks that span different levels of complexity. This shortcoming becomes especially problematic when models must solve a wide range of problems, from simple to highly intricate.

To address this gap, we draw inspiration from curriculum learning, where learners gradually progress from simpler to more challenging tasks. We propose Curriculum Demonstration Selection (CDS), a method that systematically selects demonstrations based on their complexity to ensure a balanced representation. 
It first splits the dataset into distinct difficulty levels, then retrieves demonstrations from each group. By adopting this curriculum-based approach, CDS enables LLMs to incrementally refine their understanding, which improves their performance on both simple and complex tasks. Our experiments demonstrate that CDS consistently outperforms baseline methods, particularly on more difficult problems where traditional selection techniques often fall short.

The contributions of this paper are threefold:

\emph{(i)} We introduce Curriculum Demonstration Selection (CDS), a novel method for optimizing demonstration selection in ICL, leveraging curriculum learning to enhance LLM performance.

\emph{(ii)} We empirically demonstrate the effectiveness of CDS across multiple benchmarks, showing significant improvements over existing methods.

\emph{(iii)} We provide insights into the behavior of LLMs when exposed to demonstrations of varying complexity, highlighting the potential of CDS to boost challenging problem-solving capabilities.

\section{Related Work}

\subsection{Demonstration Selection}

In-context learning (ICL) has garnered increasing attention due to its effectiveness in leveraging demonstrations for LLMs without necessitating parameter updates. A critical challenge in ICL lies in selecting optimal demonstrations, as research has shown that they can significantly influence performance \cite{dong2022survey}. Several strategies have been explored for demonstration selection. Early research use random selection \cite{brown2020language,lewkowycz2022solving} or human-crafted examples \cite{wei2022chain,kazemi-etal-2023-lambada}. While straightforward to implement, these methods often result in inconsistent performance, as some demonstrations may not provide effective guidance to the model. To mitigate this issue, \citet{liu-etal-2022-makes} proposed a retrieval-based method where examples are selected based on their semantic similarity to the query. Another variant \cite{wang2024large} uses a latent variable model to explain the effectiveness of certain demonstrations, showing that semantically similar examples can enhance generalization across tasks. Beyond semantic similarity, other strategies have emerged. Complexity-based selection \cite{fu2022complexity} emphasizes selecting demonstrations that involve more reasoning steps or higher complexity. Furthermore, \citet{zhao-etal-2024-knn} presents kNN-ICL, a method that integrates ICL with a nearest-neighbor search to simplify prompt engineering and enhance compositional task-oriented parsing. In addition, there are more advanced selection processes based on metrics such as informativeness \cite{li2023finding}, perplexity \cite{gonen-etal-2023-demystifying} and concept learning \cite{wang2024large}.

\subsection{Curriculum Learning}

The concept of curriculum learning \cite{Bengio2009CurriculumL} has inspired extensive research across numerous NLP tasks including ICL. \citet{drozdov-etal-2023-parade} prioritized harder demonstrations, estimated by Demonstration Query Likelihood (DQL). The most challenging samples, represented by low-probability queries, are selected assuming they contribute to larger updates in the learning process, emulating how gradient updates work during training. \citet{liu2024let} introduced In-Context Curriculum Learning (ICCL), which advocates ordering already selected demonstrations from simple to complex. Additionally, other approaches to demonstration selection in ICL have emerged that align with curriculum learning principles. For example, \citet{sorensen-etal-2022-information} proposed an information-theoretic framework designed to select demonstrations that maximize information gain, a strategy that can be interpreted as a form of curriculum learning where examples are chosen for their informativeness rather than difficulty. Despite these advancements, most existing methods focus primarily on selecting demonstrations similar to the test instance, according to predefined metrics, and overlook the importance of incorporating a set of demonstrations with varying complexity levels. This lack of diversity may hinder the model’s ability to generalize effectively across a broader spectrum of complexities.

\section{Methodology}

We introduce Curriculum Demonstration Selection (CDS), a novel strategy for selecting examples that draws inspiration from curriculum learning. CDS ensures that selected demonstrations cover a wide spectrum of difficulty levels, enabling LLMs to progressively learn from varied complexities within the training set.

\subsection{Problem Formulation}

Given an LLM $\theta$ and a set of $n$ training pairs ${\{x_i, y_i\}}_{i=1}^n$, our objective is to select $k$ demonstrations from this set to guide the model $\theta$ in solving a specific task $S$. We hypothesize that demonstrations with diverse complexity levels will enhance the model's performance. Therefore, CDS aims to acquire demonstrations from different difficulty levels.

\subsection{Curriculum Construction}

To construct the curriculum, we rely on human-annotated complexity measures. Complexity is assessed based on factors such as grade level, with higher grade levels indicating more challenging examples. In cases where examples share the same complexity level, we further differentiate their difficulty using human performance metrics, such as task completion rates (e.g., acceptance rates in coding tasks). The dataset is divided into $k$ distinct difficulty partitions, where $k$ is determined by the distribution of complexity scores in the dataset. This partitioning enables the construction of a curriculum that spans a broad spectrum of example complexities, ensuring that the LLM is exposed to both simpler and more complex tasks.

\subsection{Demonstration Retrieval}

Once the dataset is partitioned into $k$ difficulty levels, the CDS algorithm, outlined in Algorithm \ref{alg:cds}, selects one demonstration from each partition. This process ensures that the selected demonstrations are representative of different complexities, allowing the LLM to leverage a balanced set of examples. The retrieval process can either be similarity-based or random. For similarity-based retrieval, we use CLS embeddings from a pre-trained Transformers model to represent the sentences, following KATE \cite{liu-etal-2022-makes}. The algorithm then selects demonstrations that are most similar to the test question from each difficulty level by calculating the negative Euclidean distance. 
We then shuffle the demonstration order to prevent overfitting to specific patterns and ensure unbiasedness.

\begin{algorithm}
\caption{Curriculum Demonstration Selection (CDS)}
\label{alg:cds}
\begin{algorithmic}[1]
\State \textbf{Input: } LLM $\theta$, Training set $T$, Testing set $E$, Difficulty Measurer $D$, Retrieval function $R$, number of demonstrations $k$
\State \textbf{Output: } Selected demonstrations for each test instance

\State $\{T_i\}_{i=1}^k = D(T)$ \Comment{Partition the training set $T$ into $k$ subsets by difficulty using $D$}

\For {each $test$ instance in $E$} 
    \State $demonstration\_pool \gets$ empty list
    \For {$i = 1$ to $k$}
        \State $demonstration\_pool$.add($R(T_i)$) 
    \EndFor
    \State $demonstration\_pool$.shuffle()
    \State \textbf{$prediction = \theta(demonstration\_pool)$}
\EndFor

\end{algorithmic}
\end{algorithm}

\section{Experiments}

\begin{table*}[t!]
\centering
\scalebox{1}{
\begin{tabular}{l|cccccc |c}
\toprule
 & 
\multicolumn{6}{c|}{\textbf{MATH}} & \multirow{2}{*}{\textbf{ARC-c}} \\  
 & \textbf{Algebra} & \textbf{Geometry} & \textbf{Precalculus} & \textbf{Number Theory} & \textbf{Probability} & \textbf{Avg.} 
& \textbf{} \\ 
\midrule

\textbf{Llama 2 7B} & 
4.58$\pm$0.33 & 1.11$\pm$0.1 & 0.92$\pm$0.15 & 3.27$\pm$0.23 & 2.18$\pm$0.26 & 3.48$\pm$0.17 & 
57.37$\pm$0.43 \\

+ KATE & 
4.93$\pm$0.27 & 2.02$\pm$0.69 & \textbf{2.26$\pm$0.31} & 4.26$\pm$0.36 & 2.88$\pm$0.4 & 4.09$\pm$0.08 & 
57.54$\pm$0.43 \\ 

+ CDS (ours) & 
\textbf{5.66$\pm$0.29} & \textbf{2.16$\pm$0.26} & 2.08$\pm$0.09 & \textbf{4.57$\pm$0.68} & \textbf{3.59$\pm$0.17} & \textbf{4.62$\pm$0.24} & 
\textbf{57.99$\pm$0.26} \\ 
\midrule

\textbf{Llama 2 13B} & 
6.7$\pm$0.06 & 2.92$\pm$0.59 & 1.04$\pm$0.09 & 3.46$\pm$0.78 & 3.09$\pm$0.2 & 5.03$\pm$0.09 & 
65.02$\pm$0.12 \\ 

+ KATE & 
7.69$\pm$0.19 & 2.99$\pm$0.86 & 2.08$\pm$0.38 & 4.57$\pm$0.23 & 4.64$\pm$0.17 & 6.0$\pm$0.19 & 
66.24$\pm$0.32 \\ 

+ CDS (ours) & 
\textbf{7.94$\pm$0.1} & \textbf{3.2$\pm$0.1} & \textbf{2.32$\pm$0.17} & \textbf{5.0$\pm$0.4} & \textbf{4.92$\pm$0.5} & \textbf{6.27$\pm$0.06} & 
\textbf{66.47$\pm$0.64} \\ 
\midrule

\textbf{Llama 3 8B} & 
25.91$\pm$0.25 & 12.18$\pm$1.45 & 8.18$\pm$0.53 & 17.47$\pm$0.86 & 16.60$\pm$0.95 & 20.87$\pm$0.38 & 
80.29$\pm$0.28 \\ 

+ KATE & 
\textbf{27.14$\pm$0.18} & 13.15$\pm$0.59 & 9.65$\pm$0.23 & 16.36$\pm$0.09 & 20.39$\pm$0.78 & 22.09$\pm$0.11 & 
\textbf{81.11$\pm$0.2} \\ 

+ CDS (ours) & 
27.09$\pm$0.52 & \textbf{15.31$\pm$0.71} & \textbf{10.13$\pm$0.71} & \textbf{17.96$\pm$0.76} & \textbf{21.31$\pm$0.17} & \textbf{22.57$\pm$0.28} & 
81.0$\pm$0.15 \\ 
\midrule

\textbf{Mistral 7B} & 
12.71$\pm$0.26 & 5.15$\pm$1.0 & 4.7$\pm$0.17 & 5.74$\pm$0.76 & 7.88$\pm$0.43 & 9.9$\pm$0.03 & 
75.37$\pm$0.35 \\ 

+ KATE & 
\textbf{14.9$\pm$0.3} & \textbf{6.61$\pm$0.35} & 6.41$\pm$0.3 & 6.23$\pm$0.71 & 7.74$\pm$0.26 & 11.57$\pm$0.12 & 
76.59$\pm$0.15 \\ 

+ CDS (ours) & 
14.89$\pm$0.03 & 6.26$\pm$0.3 & \textbf{6.47$\pm$0.09} & \textbf{6.42$\pm$0.87} & \textbf{8.65$\pm$0.52} & \textbf{11.64$\pm$0.02} & 
\textbf{76.71$\pm$0.25} \\ 
\midrule

\textbf{Qwen 7B} & 
13.75$\pm$0.14 & 5.43$\pm$1.02 & 5.56$\pm$1.27 & 6.54$\pm$0.71 & 8.79$\pm$0.1 & 10.81$\pm$0.21 & 
48.66$\pm$0.79 \\ 

+ KATE & 
15.25$\pm$0.25 & 6.96$\pm$0.43 & \textbf{6.41$\pm$0.79} & 7.9$\pm$0.75 & 9.14$\pm$0.53 & 12.12$\pm$0.25 & 
44.62$\pm$0.24 \\ 

+ CDS (ours) & 
\textbf{15.54$\pm$0.26} & \textbf{7.24$\pm$0.94} & 5.92$\pm$0.46 & \textbf{8.64$\pm$1.06} & \textbf{9.7$\pm$0.6} & \textbf{12.39$\pm$0.04} & 
\textbf{49.74$\pm$0.48} \\ 
\bottomrule
\end{tabular}
}
\caption{Result on MATH and ARC-c dataset. Numbers in bold indicate the best scores among all methods.}
\label{reasoning}
\end{table*}

\subsection{Experimental Setup}
\label{exset}
\subsubsection{Datasets} 

\paragraph{\textbf{$\bullet$ Math reasoning}} Mathematical reasoning plays a crucial role in evaluating LLMs' performance. Numerous studies have focused on leveraging LLMs on math-related tasks such as mathematical question generation \cite{Kurdi2019ASR, 10.1145/3605098.3636030} and problem solving \cite{zhou2023solving, gou2023tora, chen2024empirical}. In this study, we utilize the MATH dataset \cite{hendrycks2021measuring}, a highly challenging benchmark consisting of 7,500 training examples and 5,000 testing examples, spanning topics such as prealgebra, algebra, number theory, counting and probability, geometry, intermediate algebra, and precalculus. The MATH dataset, sourced from competitions such as the AMC 10, AMC 12, and AIME, categorizes problems into five distinct complexity levels. We leverage this metadata to construct the curriculum for our CDS method, ensuring that demonstrations reflect a range of difficulty levels.

\paragraph{\textbf{$\bullet$ Commonsense reasoning}} Commonsense reasoning, essential for natural language understanding, tests a model’s ability to infer everyday scenarios by integrating observational and prior knowledge \cite{talmor-etal-2019-commonsenseqa, sakaguchi2021winogrande}. To evaluate LLM performance in this area, we use the ARC-Challenge dataset \cite{clark2018think}, which contains 2,590 multiple-choice science questions targeting students from grade 3 to grade 9. For CDS, we rank the difficulty by the grade level data, ensuring a balanced representation across complexity levels.

\paragraph{\textbf{$\bullet$ Code generation}} These tasks have become pivotal in assessing LLMs’ capabilities, covering areas from code correction \cite{sobania2023analysis, zhang2023critical} to code summarization \cite{sun2023automatic} and efficiency optimization \cite{du2024mercury}. For this task, we employ the Mercury dataset, a specialized benchmark designed for evaluating code generation correctness and efficiency. Mercury consists of 256 evaluation tasks sourced from public programming problems on Leetcode\footnote{https://leetcode.com/problemset/algorithms/}, categorized into Easy, Medium, and Hard difficulty levels. To further refine difficulty classification, we use acceptance rates as an additional metric, with lower acceptance rates indicating higher complexity within the same difficulty category.

\subsubsection{Models}

For the math and commonsense reasoning tasks, we test five widely used open-source LLMs: Llama-2 (7B and 13B) \cite{touvron2023llama}, Llama-3 (8B) \cite{dubey2024llama}, Mistral-7B \cite{jiang2023mistral}, and Qwen-7B \cite{bai2023qwen}. For the code generation task, we follow the experimental settings of Mercury \cite{du2024mercury}, evaluating four code-specialized LLMs: CodeLlama (7B and 13B) \cite{roziere2023code}, StarCoder (3B) \cite{li2023starcoder}, and DeepSeek-Coder (6.7B) \cite{guo2024deepseek}, with model sizes ranging from 3B to 13B parameters.

\subsubsection{Implementation Details}

We employ greedy decoding across all models for math and commonsense reasoning tasks. Prompts are constructed using the Chain-of-Thought (CoT) technique \cite{wei2022chain}, which encourages step-by-step reasoning before arriving at the final answer. The final prediction is extracted from the last line of the model’s output. In the Mercury dataset, we adhere to the settings of \citet{du2024mercury}, with a decoding temperature of 0.2. All experiments use $k=5$ demonstrations, and each model is evaluated using three different random seeds on the test sets. Example prompts in our experiments are provided in \Cref{prompts}.

\subsubsection{Baselines}

We compare CDS with two baseline methods:

\textbf{$\bullet$ Uniform:} We uniformly select $k$ demonstrations from training set $T$ for each test example.

\textbf{$\bullet$ Similarity (KATE \cite{liu-etal-2022-makes}):} This method utilizes RoBERTa-Large \cite{liu2019roberta} to obtain CLS embeddings, followed by cosine similarity computation between candidate demonstrations and the test examples. The $k$ most similar demonstrations are then selected based on negative Euclidean distance. 

\subsubsection{Evaluation metrics}

For the math and commonsense reasoning tasks, accuracy is used as the evaluation metric. A prediction is considered correct if it exactly matches the ground truth, either as a value (MATH) or a selected option (ARC-Challenge). For the code generation task, we use two key metrics from the Mercury dataset. The first is Functional Correctness (Pass), which evaluates whether the generated code successfully passes all test cases associated with a given task. The Pass score represents the percentage of tasks for which the code is fully functional. The second metric, Beyond, evaluates not only the correctness of the generated code but also its efficiency. This metric ranks the runtime performance of the model-generated code against historical solutions, providing a percentile score that reflects how efficiently the code executes.

\subsection{Results}

\subsubsection{LLM Reasoning}

We compared five widely-used open-source LLMs on mathematical and commonsense reasoning tasks, where the results clearly demonstrate the superiority of CDS over other methods, as seen in \Cref{reasoning}. In the mathematical reasoning benchmark, CDS consistently outperforms both random selection and the KATE baseline. Across all mathematical topics in the MATH dataset, CDS delivers stronger performance, with algebra and number theory benefiting the most, while geometry and precalculus show moderate improvements. Notably, probability problem solve rates increase substantially with CDS. Larger models, particularly Llama 2 13B, experience the greatest performance gains, underscoring the scalability of CDS in enhancing model capabilities, in line with the Scaling Laws proposed by Kaplan et al. \cite{Kaplan2020ScalingLF}.

Similarly, on the ARC-c benchmark, CDS outperforms both random selection and KATE. While KATE provides some improvement over random selection, its performance is inconsistent across tasks. For example, KATE underperforms the random baseline by approximately 9\% with Qwen 7B. In contrast, CDS exhibits greater stability, consistently yielding superior results across the majority of cases. This highlights CDS as an effective and reliable method for in-context learning in LLMs.

\subsubsection{Code Generation}

CDS also demonstrates significant improvements in code generation tasks, as evaluated on the Mercury dataset in \Cref{tab:code_generation}. Specifically, CDS significantly outperforms the random baseline in both the Pass and Beyond metrics for all LLMs. Furthermore, CDS consistently provides improvement over the KATE baseline on the Pass metric across all models and shows comparable results on the Beyond metric. These results highlight the effectiveness of CDS on the code generation task.

\begin{table}
\centering
\begin{tabular}{l|c|c}
\toprule
\textbf{Model} & \textbf{Pass} & \textbf{Beyond} \\ 
\midrule
\textbf{CodeLlama-7b }        & 28.12$\pm$0.68                      & 21.75$\pm$0.65                     \\ 
+ KATE & 36.85$\pm$0.9 & \textbf{28.34$\pm$1.8} \\ 
+ CDS (ours)  & \textbf{36.98$\pm$1.93}   &  28.1$\pm$2.0 \\

\midrule
\textbf{CodeLlama-13b}         & 33.98$\pm$1.03                    &   26.86$\pm$0.9              \\ 
+ KATE & 39.45$\pm$0.01 &  32.24$\pm$0.27\\
+ CDS (ours)                   & \textbf{41.54$\pm$0.45}                      & \textbf{33.18$\pm$0.76}                  \\ 
\midrule
\textbf{deepseek-coder-6.7b}         & 43.36$\pm$0.68                    & 33.95$\pm$0.52                   \\ 
+ KATE &  64.19$\pm$0.9 &  \textbf{52.24$\pm$0.11} \\ 
+ CDS (ours)                   &\textbf{64.71$\pm$0.9}                    & 51.51$\pm$1.38                      \\ 

\midrule
\textbf{starcoder2-3b}         & 50.26$\pm$1.13                 & 39.51$\pm$0.67  \\ 
+ KATE &  57.42$\pm$1.35 & 46.58$\pm$3.28 \\ 
+ CDS (ours)                   & \textbf{58.98$\pm$0.01}        & \textbf{47.15$\pm$1.06} \\
\bottomrule
\end{tabular}
\caption{Results on Mercury. Numbers in bold indicate the best scores among all methods.}
\label{tab:code_generation}
\end{table}

\subsubsection{Random retrieval v.s. Similarity retrieval}

\begin{figure}
    \centering
    \includegraphics[width=\linewidth]{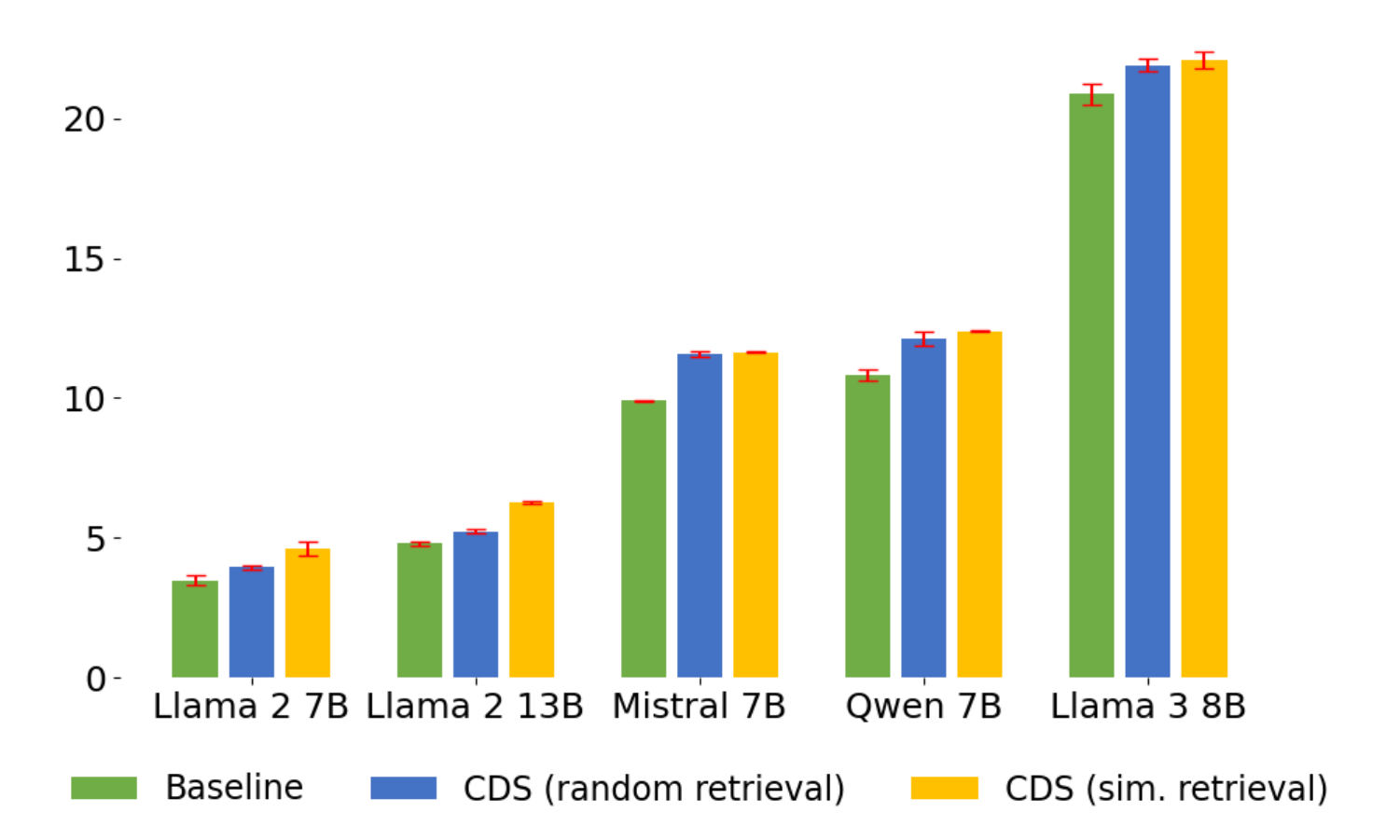}
    \caption{Comparison of CDS with random demonstration retrieval and with similarity retrieval on MATH benchmark across five LLMs.}
    \label{math_avg_comparison}
\end{figure}

We examined the effects of different retrieval functions, namely random retrieval and similarity retrieval, on CDS performance. As shown in \Cref{math_avg_comparison}, both retrieval methods outperform the random selection baseline on the MATH dataset, with similarity-based retrieval consistently outperforming random retrieval. This result suggests that similarity-based retrieval enhances the effectiveness of CDS, providing better demonstration selection.

\subsubsection{Solve rates on harder problems}

\begin{figure}
\includegraphics[width=0.99\columnwidth]{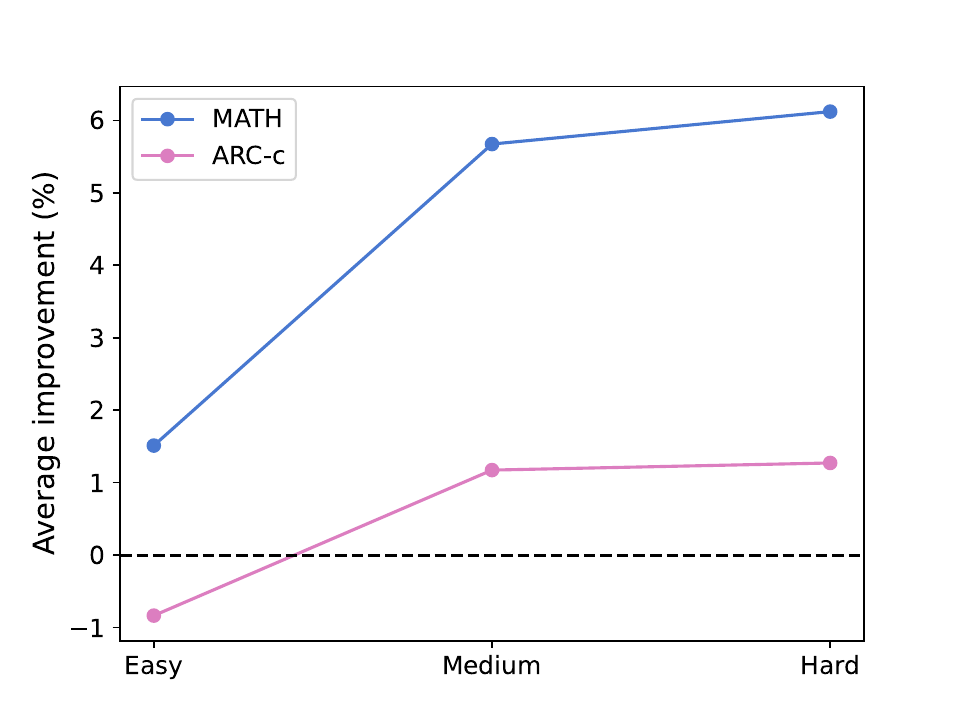}
\caption{Average improvement across five models in three difficulty levels.}
\label{reasoning_trend}
\end{figure}

Both the MATH and ARC-c datasets show a clear positive trend, with harder problems benefiting the most from CDS, as illustrated in Figure \ref{reasoning_trend}. On the MATH dataset, the performance improvement grows from 2\% on easier tasks to 6\% on harder ones. The ARC-c shows a similar, albeit more modest, trend, with improvements ranging from negative values on easy tasks to approximately 1\% on medium and hard tasks. This highlights the remarkable efficacy of CDS in addressing more complex problems, likely due to its strategy of selecting demonstrations across a wide range of difficulty levels, helping LLMs adapt better to challenging tasks.

\subsubsection{Effect of demonstrations’ order}

As demonstrated in \Cref{notshuffle}, ordering the demonstrations from easy to hard (E2H) has no significant effect on performance compared to random shuffling, indicating that the order in which demonstrations are presented does not impact the overall effectiveness of CDS. This finding aligns with previous research by Wang et al. \cite{wang2024large}, further supporting the robustness of CDS in enhancing LLM performance, regardless of demonstration order.

\begin{table}[H]
\setlength{\tabcolsep}{4pt}
\renewcommand{\arraystretch}{0.9}
\resizebox{0.99\columnwidth}{!}{
    \begin{tabular}{c|ccccccc}
    \toprule
    & \textbf{Alg.} & \textbf{Geo.} & \textbf{Cal.} & \textbf{Num. Theory} & \textbf{Prob.} & \textbf{Avg.}  \\
    \midrule
    E2H & 5.47 & 2.51 & 2.20 & 4.63 & 3.37 & 4.54 \\ 
    Rand. & 5.66$\pm$0.29 & 2.16$\pm$0.26 & 2.08$\pm$0.09 & 4.57$\pm$0.68 & 3.59$\pm$0.17 & 4.62$\pm$0.24 \\
    \bottomrule
    \end{tabular}}
\caption{Accuracy of our method with (E2H) and without (Rand.) reordering.  Numbers are obtained with the Llama-2 7B model on MATH dataset.}
\label{notshuffle}
\end{table}
\section{Conclusions}
In this paper, we introduced Curriculum Demonstration Selection (CDS), a novel approach aimed to enhance the performance of LLMs in ICL. By leveraging the principles of curriculum learning, CDS effectively organizes demonstrations according to their complexity, allowing LLMs to learn from simpler to more complex tasks. Our extensive experiments across three benchmarks — mathematical reasoning, commonsense reasoning, and code generation — consistently demonstrated that CDS outperforms traditional methods, including both random selection and similarity-based approaches like KATE. Additionally, CDS shows significant potential in solving more complex and challenging problems, highlighting its robustness in optimizing demonstration selection for ICL. Overall, CDS presents a significant step forward in optimizing demonstration selection for ICL, providing a more structured and effective method that enhances both the accuracy and efficiency of LLMs. This work opens new avenues for improving LLM in numerous problem-solving tasks.

\section{Limitations}

Our work does have some limitations. First, we employed a fixed five-shot setting in all experiments, providing the model with five demonstrations per task. While this is a common approach in in-context learning, it may not fully capture the potential of CDS when applied to different shot settings. The decision to use five demonstrations balanced computational costs and performance consistency across benchmarks. Future research could explore how varying the number of demonstrations influences CDS effectiveness, particularly for more complex tasks that may require additional context for optimal performance.

Second, the curriculum structure in CDS relies on predefined metadata, such as grade levels in math reasoning, to determine task difficulty. While this allows for efficient curriculum construction, it assumes that such metadata is available and accurately reflects task complexity. In some datasets or domains, difficulty rankings may not be available or may not accurately represent the challenges posed by the tasks. In these cases, CDS would require alternative methods for estimating task complexity, such as model-based estimates or task-specific heuristics, to maintain its effectiveness.

Lastly, while our evaluation focused on three benchmarks, namely mathematical reasoning, commonsense reasoning, and code generation, the generalizability of CDS to other types of tasks remains an open question. Future work could expand the scope of CDS evaluation to include a broader range of benchmarks, helping to identify its strengths and limitations across various domains and task types. This will provide deeper insights into the applicability of CDS and its potential to further improve LLM performance in diverse contexts.

\section*{Acknowledgements}
This research/project is supported by the National Research Foundation, Singapore under its AI Singapore Programme (AISG Award No: AISG2-TC2022-005).

%% file: appendiz.tex
\appendix

\section{Prompts used for few-shot learning}
\label{prompts}

\begin{table}[H]
\centering
\footnotesize
\tcbset{
    tabularbox/.style={colback=white, colframe=black, boxrule=0.1pt, arc=5mm}
}
\begin{tcolorbox}[tabularbox]
\begin{tabular}{p{0.9\textwidth}}
\raggedright
\#\#\# Question: A string \(s\) is called good if there are no two different characters in \(s\) that have the same frequency. Given a string \(s\), return the minimum number of characters you need to delete to make \(s\) good. The frequency of a character in a string is the number of times it appears in the string. For example, in the string "aab", the frequency of 'a' is 2, while the frequency of 'b' is 1.

\textbackslash{}textbf\{Example 1:\}
\textbackslash{}textbf\{Input:\} \textbackslash{}(s = \textbackslash{}text\{\"aab\"\}\textbackslash{}) \textbackslash{}\textbackslash{}
\textbackslash{}textbf\{Output:\} 0 \textbackslash{}\textbackslash{}
\textbackslash{}textbf\{Explanation:\} \textbackslash{}(s\textbackslash{}) is already good.

\textbackslash{}textbf\{Example 2:\}

\textbackslash{}textbf\{Input:\} \textbackslash{}(s = \textbackslash{}textbf\{\"aaabbbcc\"\}\textbackslash{}) \textbackslash{}\textbackslash{}
\textbackslash{}textbf{Output:} 2 \textbackslash{}\textbackslash{}
\textbackslash{}textbf{Explanation:} You can delete two 'b's resulting in the good string "aaabcc". Another way is to delete one 'b' and one 'c', resulting in the good string "aaabbc".

\textbackslash{}textbf\{Example 3:\}

\textbackslash{}textbf\{Input:\}(s = \textbackslash{}text\{\"ceabaacb\"\}) \\
\textbackslash{}textbf\{Output:\} 2 \\
\textbackslash{}textbf\{Explanation:\} You can delete both 'c's, resulting in the good string "eabaab". Note that we only care about characters that are still in the string at the end (i.e., frequency of 0 is ignored).

\textbackslash{}textbf\{Constraints:\}
\textbackslash{}begin\{itemize\}
    \textbackslash{}item \textbackslash{}(1 \textbackslash{}leq s.length \textbackslash{}leq $10^5$
    \textbackslash{}item \textbackslash{}(s\textbackslash{}) contains only lowercase English letters.
\textbackslash{}end\{itemize\}

\#\#\# Code Prompt: \begin{verbatim}
class Solution(object):
    def minDeletions(self, s):
\end{verbatim}
\#\#\# Code Completion:  \begin{verbatim}
import heapq
class Solution(object):
    def minDeletions(self, s):
        lst = list(s)
        counter = {}
        for char in lst:
            if char not in counter:
                counter[char] = 0
            counter[char] += 1
        q = []
        ans = 0
        for char in counter:
            heapq.heappush(q, (-counter[char], char))
        del counter
        while q:
            curr_ct, curr = heapq.heappop(q)
            if q:
                if q[0][0] == curr_ct:
                    curr_ct += 1
                    ans += 1
                    if curr_ct < 0:
                        heapq.heappush(q, (curr_ct, curr))
        return ans
\end{verbatim}
\#\#\# Question: <<Q2>>\\
\#\#\# Code Prompt: <<S2>>\\
\#\#\# Code Completion: <<E2>> \\ 
\#\#\# Question: <<Q3>>\\
\#\#\# Code Prompt: <<S3>>\\
\#\#\# Code Completion: <<E3>> \\ 
\#\#\# Question: <<Q4>>\\
\#\#\# Code Prompt: <<S4>>\\
\#\#\# Code Completion: <<E4>> \\ 
\#\#\# Question: <<Q5>>\\
\#\#\# Code Prompt: <<S5>>\\
\#\#\# Code Completion: <<E5>> \\ 
\#\#\# Question: <<Test Question>>\\
\#\#\# Code Prompt: <<Test Code Prompt>>\\
\end{tabular}
\end{tcolorbox}
\caption{Example of the prompt used in the experiments on the Mercury dataset.}
\label{tab:prompt_example_mer}
\end{table}

\begin{table}[H]
\centering
\footnotesize
\tcbset{
    tabularbox/.style={colback=white, colframe=black, boxrule=0.1pt, arc=5mm}
}
\begin{tcolorbox}[tabularbox]
\begin{tabular}{p{0.9\textwidth}}
\raggedright
\#\#\# Question: The least common multiple of two integers is 36 and 6 is their greatest common divisor. What is the product of the two numbers?\\
\#\#\# Solution: Let \$a\$ and \$b\$ be the two integers. We can use the identity \$\textbackslash{}gcd(a,b) \textbackslash{}cdot  \textbackslash{}mathop{ \textbackslash{}text{lcm}}[a,b] = ab\$. Substituting gives that the answer is \$36  \textbackslash{}cdot 6 =  \textbackslash{}boxed\{216\}\$.\\
\#\#\# Extracted Answer: 216 \\ 
\#\#\# Question: In an office at various times during the day, the boss gives the secretary a letter to type, each time putting the letter on top of the pile in the secretary's in-box. When there is time, the secretary takes the top letter off the pile and types it. There are nine letters to be typed during the day, and the boss delivers them in the order \$1, 2, 3, 4, 5, 6, 7, 8, 9\$. While leaving for lunch, the secretary tells a colleague that letter $8$ has already been typed, but says nothing else about the morning's typing. The colleague wonders which of the nine letters remain to be typed after lunch and in what order they will be typed. Based upon the above information, how many such after-lunch typing orders are possible? (That there are no letters left to be typed is one of the possibilities.) Re-stating the problem for clarity, let \$S\$ be a set arranged in increasing order. At any time an element can be appended to the end of \$S\$, or the last element of \$S\$ can be removed. The question asks for the number of different orders in which the all of the remaining elements of \$S\$ can be removed, given that \$8\$ had been removed already.\\
\#\#\# Solution: Since \$8\$ had already been added to the pile, the numbers \$1 \textbackslash{}ldots 7\$ had already been added at some time to the pile; $9$ might or might not have been added yet. So currently \$S\$ is a subset of \$\textbackslash{}\{1, 2, \textbackslash{}ldots 7\textbackslash{}\}\$, possibly with \$9\$ at the end. Given that \$S\$ has \$k\$ elements, there are \$k+1\$ intervals for \$9\$ to be inserted, or \$9\$ might have already been placed, giving \$k+2\$ different possibilities. Thus, the answer is \$\textbackslash{}sum\_\{k=0\}\^\{7\} \textbackslash{}7 \textbackslash{}choose k\}(k+2)\$ \$= 1 \textbackslash{}cdot 2 + 7 \textbackslash{}cdot 3 + 21 \textbackslash{}cdot 4 + 35 \textbackslash{}cdot 5 + 35 \textbackslash{}cdot 6 + 21 \textbackslash{}cdot 7 + 7 \textbackslash{}cdot 8 + 1 \textbackslash{}cdot 9\$ \$= \textbackslash{}boxed\{704\}\$.
\\
\#\#\# Extracted Answer: 704 \\
\#\#\# Question: <<Q3>>\\
\#\#\# Solution: <<S3>>\\
\#\#\# Extracted Answer: <<E3>> \\
\#\#\# Question: <<Q4>>\\
\#\#\# Solution: <<S4>>\\
\#\#\# Extracted Answer: <<E4>> \\
\#\#\# Question: <<Q5>>\\
\#\#\# Solution: <<S5>>\\
\#\#\# Extracted Answer: <<E5>> \\
\#\#\# Question: <<Test Question>>\\
\end{tabular}
\end{tcolorbox}
\caption{Example of the prompt used in the experiments on the MATH dataset.}
\label{tab:prompt_example}
\end{table}

\label{appendix}